\documentclass{article}
\usepackage[T1]{fontenc}
\usepackage{iclr2027_conference,times}
\usepackage{amsmath,amssymb,amsthm}
\usepackage{graphicx,booktabs,multirow,array}
\usepackage{xcolor}
\usepackage{wrapfig,adjustbox,enumitem,pifont,colortbl,placeins,float}
\usepackage{hyperref}
\usepackage{fontawesome5}
\definecolor{ourspeach}{HTML}{F8E6D8}

\usepackage{hyperref}
\usepackage{url}

\title{Prefill-Free Cross-Family KV Cache Transfer for Heterogeneous Multi-Agent LLMs}

\author{
Vincent-Daniel Yun\textsuperscript{1,*},
Woosang Lim\textsuperscript{2,*},
Haneul Yoo\textsuperscript{3},
Sungjoo Yoo\textsuperscript{2},\\
\textbf{
Murali Annavaram\textsuperscript{1},
Sai Praneeth Karimireddy\textsuperscript{1,\dag}
}\\
1. University of Southern California, USA\\
2. Seoul National University, Republic of Korea\\
3. New York University, USA\\
\texttt{\{yunjuyou, annavara, karimire\}@usc.edu}\\
\texttt{ftyg656512@snu.ac.kr, sungjoo.yoo@gmail.com}\\
\texttt{haneul.yoo@nyu.edu}\\
\textsuperscript{*}Equal Contribution.
\textsuperscript{\dag}Corresponding Author.
}

\iclrpreprintcopy

\begin{document}

\maketitle

\vspace{-0.5cm}
\begin{center}
\href{https://github.com/daniel-eai/Prefill-Free-Multi-Agent-LLMs}
{\faGithub\ GitHub}
\end{center}

\begin{abstract}
Recent multi-agent LLM systems increasingly combine heterogeneous models for specialized agent roles. However, text-based communication requires each receiver to prefill shared context already processed by the sender. Reusing the sender's key-value (KV) cache avoids this redundancy, but prefill-free transfer across model families must handle differences in tokenization, model depth, and KV representations. To address these issues, we propose \textit{HeteroFold}, a prefill-free cross-family KV cache transfer method that keeps both the sender and receiver frozen. HeteroFold aligns model structures, maps the sender cache into the receiver space, and calibrates it to preserve receiver behavior. Across six transfer directions, HeteroFold achieves the best cache-transfer performance on all four long-context benchmarks and most short-context settings. It also matches text-based communication on the multi-agent benchmark. At 32K context length, Llama-3.1-8B$\rightarrow$Ministral-3-14B transfer is $10.7\times$ faster than Native Prefill and $1.18$--$1.47\times$ faster than the state-of-the-art prefill-free baselines, Dense Latent and KV Ridge. These results show that HeteroFold enables efficient cross-family KV reuse without receiver prefill.
\end{abstract}

\section{Introduction}

\begin{figure}[h]
\centering
\includegraphics[width=\textwidth]{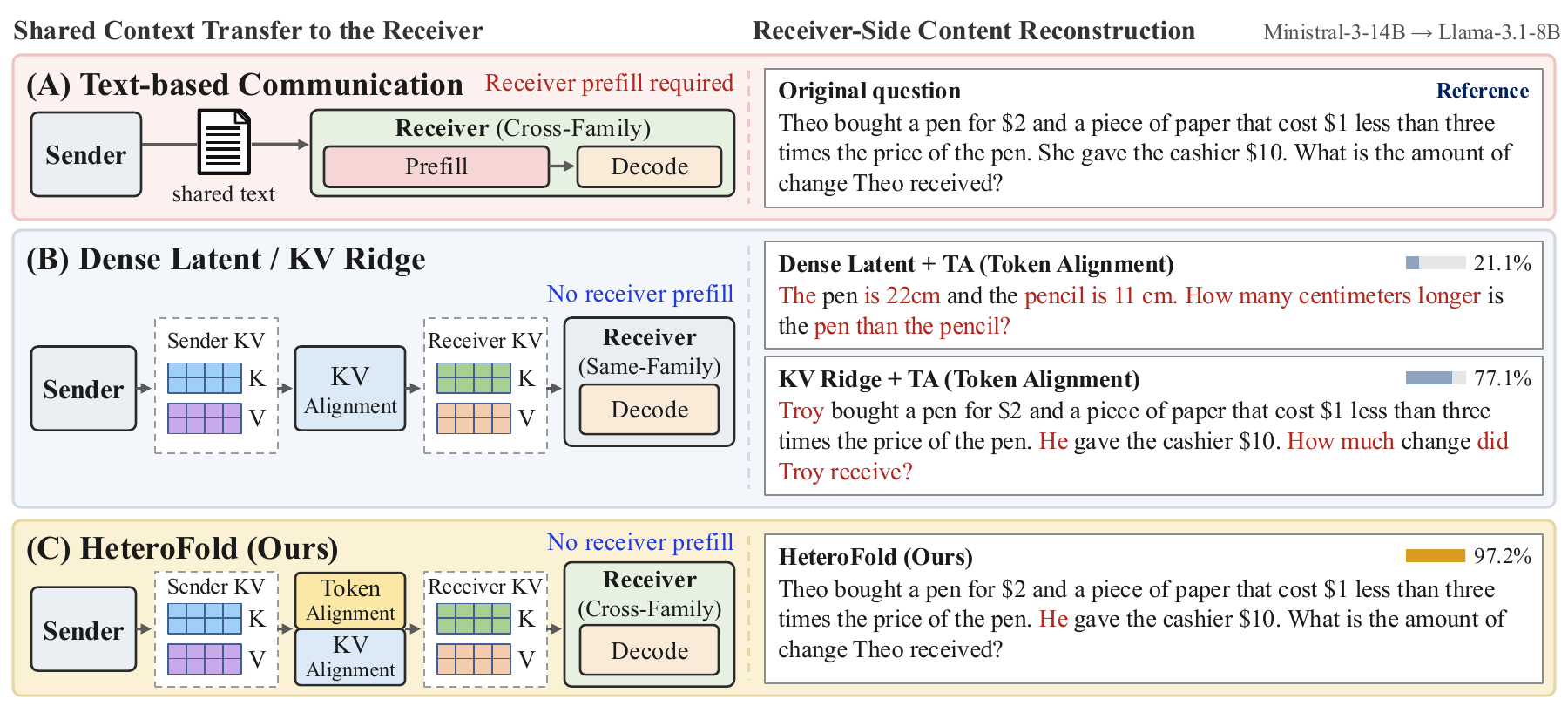}
\vspace{-0.5cm}
\caption{Comparison of communication settings and receiver-side question
reconstruction. HeteroFold enables prefill-free cross-family KV cache transfer.
For reconstruction, the sender prefills the question, transfers its KV cache,
and the receiver is prompted to restate the question without the original text.
Panel (B) shows the original same-family setting of prior cache-transfer
methods; reconstruction uses TA for cross-family Ministral-3-14B$\rightarrow$Llama-3.1-8B transfer on GSM8K. Red text marks differences from the reference. Percentages report ordered overlap of this prompt.}
\label{fig:communication-hero}
\end{figure}

Large language models (LLMs) are increasingly used as collaborating agents in
multi-agent systems (MAS), which decompose tasks across specialized
roles~\citep{hong2023metagpt}, coordinate through
conversation~\citep{wu2023autogen}, and make decisions through
consensus~\citep{chen2023consensus,lee2026robust}.
We refer to systems that combine models from different families or scales as
\emph{heterogeneous multi-agent systems}. These systems can assign models to
roles based on their capabilities and computational costs~\citep{wang2025mixture};
for example, X-MAS~\citep{ye2025x} uses different model families for solver,
evaluator, and aggregator roles.
However, heterogeneous agents often exchange shared context. Text-based
communication requires each receiver to prefill it again and rebuild its KV
cache~\citep{woo2026prefillshare}, adding redundant computation and latency as
context length grows~\citep{linearkv2026}.

\begin{figure}[t]
\centering
\includegraphics[width=\textwidth]{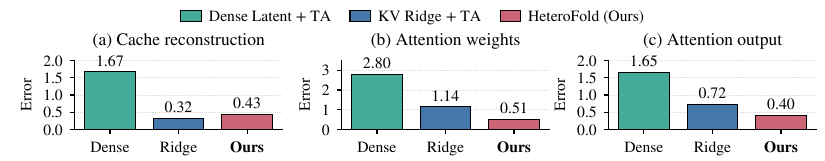}
\vspace{-0.6cm}
\caption{Cache and attention errors after transfer, averaged over all six
transfer directions on Qasper. Panels (a,c) report normalized squared error
for K/V reconstruction and attention output; (b) reports attention-weight KL
divergence. Both baselines use Token Alignment (TA); lower is better.}
\label{fig:functional-analysis}
\end{figure}

Cross-family KV reuse can remove repeated receiver prefill, but must handle
mismatches in tokenization, model depth, and KV representations.
Figure~\ref{fig:communication-hero} illustrates how these mismatches can affect
receiver-side text reconstruction, while Figure~\ref{fig:functional-analysis} shows that cache
reconstruction error alone does not reflect receiver behavior: KV
Ridge~\citep{linearkv2026} achieves lower reconstruction error than HeteroFold,
yet distorts receiver attention. Therefore, effective transfer must preserve not just the cache values, but the downstream receiver computation they induce.

We propose \textbf{HeteroFold}, to our knowledge the first receiver-prefill-free
cross-family KV cache transfer method with explicit cross-tokenizer alignment.
HeteroFold addresses \textbf{heterogeneity} across model families by aligning tokens and
layers, mapping K/V features to receiver statistics, and calibrating the
transferred cache against native receiver attention and outputs. The learned
corrections are \textbf{folded} into fixed affine maps, keeping both models frozen and
enabling direct decoding without receiver prefill or an additional inference
module.

Across six transfer directions among Llama, Qwen, and Ministral, HeteroFold
outperforms the evaluated cache-transfer baselines on all long-context
benchmarks and in most short-context benchmarks.
On \textsc{HiddenBench}, it achieves group decision accuracy comparable to
text-based communication without receiver prefill. At 32K context length,
Llama-3.1-8B$\rightarrow$Ministral-3-14B transfer is $10.7\times$ faster than
Native Prefill and $1.18$--$1.47\times$ faster than the state-of-the-art
prefill-free baselines, KV Ridge and Dense Latent.

Our contributions are:
\begin{itemize}[leftmargin=*]

\item We identify tokenizer mismatch as a key obstacle to prefill-free
cross-family KV cache reuse and introduce \textit{Token Alignment (TA)}, which
establishes sender--receiver correspondence through shared character
boundaries.

\item We develop a cross-family K/V mapping that handles differences in model
depth, KV structure, and feature distributions. It combines multi-layer sender
features, cross-head mixing, and Recolor to construct receiver-compatible K/V
states while keeping both language models frozen.

\item We show that KV reconstruction alone does not preserve receiver behavior
and introduce receiver-aware calibration that directly matches attention
patterns and outputs. The learned corrections fold into fixed affine maps,
enabling direct decoding without receiver prefill.

\end{itemize}

\section{Related Work}
\textbf{Cross-family communication.}
Standard text communication, denoted TextMas following \citet{zou2025latent}, supports cross-family exchange but requires each receiver
to prefill the exchanged text and rebuild its KV cache. C2C~\citep{fu2026c2c}
supports cross-model cache communication, but requires a prompt cache generated
by the receiver. Therefore, the receiver must process the shared prompt before
cache transfer, so C2C does not eliminate receiver prefill.
\textsc{RecursiveMAS}~\citep{zou2026recursive} communicates across
heterogeneous models through hidden representations, but the transferred
information is still processed by the receiver's layers rather than provided as
a decode-ready KV cache. 

\noindent\textbf{Prefill-free communication.}
\textsc{LatentMAS}~\citep{zou2025latent} avoids standard text communication by
sharing latent states and layer-wise caches, but assumes cache compatibility
between the communicating models. This assumption does not directly cover
models that differ in tokenization, depth, or KV structure. Dense
Latent~\citep{chen2026dense} and KV Ridge~\citep{linearkv2026} map sender cache
states into receiver cache states and avoid receiver prefill. However, their reported
evaluations remain within model families with compatible
tokenization. As a result, they do not address the token correspondence,
layer alignment, and KV representation mismatches introduced by cross-family
transfer.

Previous research has explored partial solutions for KV-cache transfer: cross-family
communication still requires receiver-side processing, while prefill-free cache
transfer is applicable when sender and receiver structures are compatible. 
In contrast, HeteroFold addresses both by aligning different
tokenizers and model depths, mapping sender K/V features into the receiver
space, and preserving receiver behavior. This enables direct cross-family KV
cache transfer without receiver prefill.

\begin{figure}[t]
  \centering
  \includegraphics[width=\linewidth]{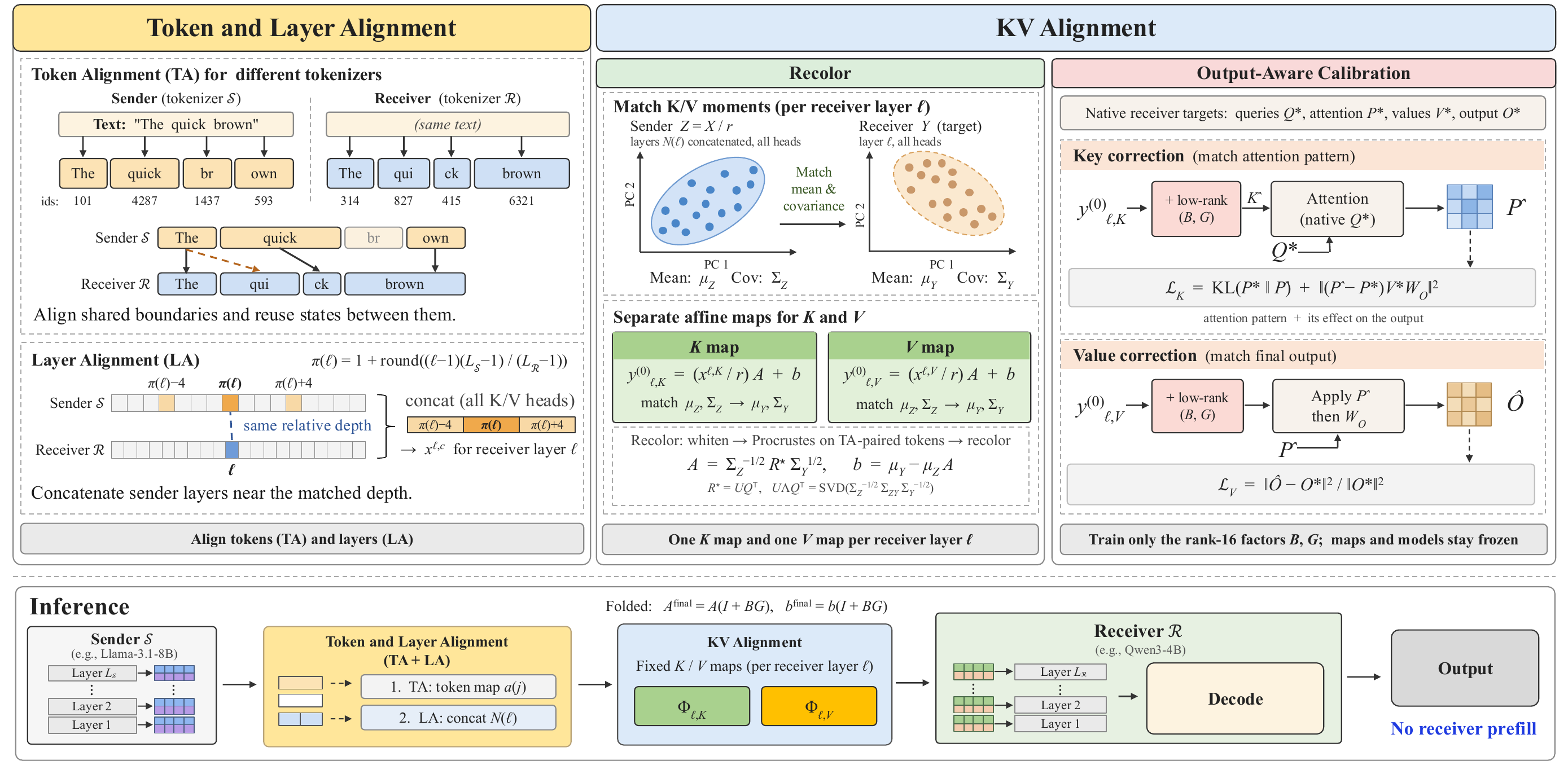}
  \caption{HeteroFold aligns tokens and layers, initializes separate K/V maps, and calibrates receiver attention and outputs. Corrections fold into affine maps; both models remain frozen.}
  \label{fig:method-overview}
\end{figure}

\section{Method: HeteroFold}
\label{sec:method}

Let $\mathcal S$ and $\mathcal R$ denote frozen sender and receiver models
with $L_{\mathcal S}$ and $L_{\mathcal R}$ layers. Given context $x$,
HeteroFold maps $\mathcal C_{\mathcal S}(x)$ to a receiver-compatible cache
$\widehat{\mathcal C}_{\mathcal R}(x)$ without receiver prefill. 
We map keys before key normalization and RoPE~\citep{su2024roformer} and
values after projection; the receiver then applies its native key normalization
and RoPE. HeteroFold consists of token and layer alignment, Recolor KV
mapping, and output-aware calibration (Figure~\ref{fig:method-overview}).

\subsection{Token and Layer Alignment}
\label{sec:ta}\label{sec:alignment}

\paragraph{Token Alignment (TA).}
TA aligns tokens through shared character-end boundaries.
For an unmatched receiver boundary, we use the sender state at the latest
preceding shared boundary, or the first sender state before the first match.
In many-to-one cases, sender states are not averaged.
After K/V mapping, the receiver applies RoPE using its own position indices.
The following example illustrates different tokenizations of
``The unbelievable result.'':
\begin{center}
\footnotesize
\begin{tabular}{@{}lll@{}}
\toprule
\textbf{Model} & \textbf{Tokens} & \textbf{End positions} \\
\midrule
Qwen3
& \texttt{The $\mid$ unbelievable $\mid$ result $\mid$ .}
& $3,\,16,\,23,\,24$ \\
Ministral-3
& \texttt{The $\mid$ unbel $\mid$ iev $\mid$ able $\mid$ result $\mid$ .}
& $3,\,9,\,12,\,16,\,23,\,24$ \\
\bottomrule
\end{tabular}
\end{center}

For Qwen3$\rightarrow$Ministral-3, receiver boundaries at characters $9$ and
$12$ reuse the Qwen state ending at character $3$. In the reverse direction,
the Qwen token ending at character $16$ uses the Ministral state ending at the
same boundary, matching the same causal prefix endpoint rather than averaging
intermediate states.

\begin{figure}[t]
  \centering
  \includegraphics[width=\linewidth]{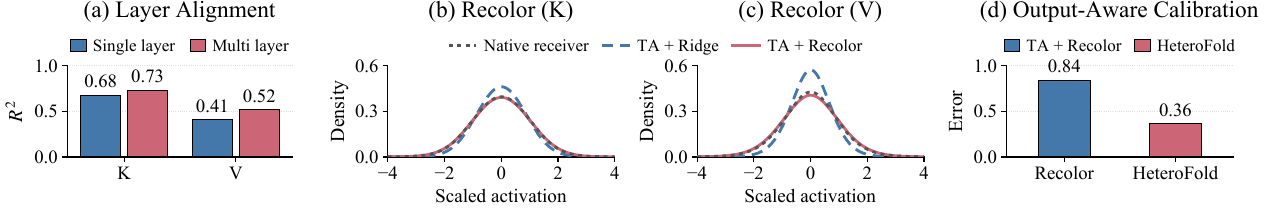}
  \vspace{-0.7cm}
  \caption{Motivation for HeteroFold components on 100 held-out HotpotQA prompts
    for Ministral-3-14B$\rightarrow$Llama-3.1-8B. (a) Layer Alignment:
    multi-layer inputs yield higher $R^2$ for predicting native receiver K/V than
    a single layer. (b,c) Recolor: K/V activation densities at receiver layer 16,
    standardized using native per-feature statistics. (d) Output-Aware Calibration:
    calibration reduces the attention-output error after Recolor.}
  \label{fig:design-motivation}
\end{figure}

\paragraph{Layer Alignment (LA).}
Across model families, receiver-relevant information may be distributed across
multiple sender layers. Figure~\ref{fig:design-motivation}(a) shows that
aggregating neighboring sender layers improves prediction of native receiver
K/V representations over a single layer. We use proportional depth
alignment with three sender layers at offsets $(-4,0,+4)$ around the matched
layer:
\begin{equation}
  \pi(\ell)=1+\operatorname{round}\left(
  (\ell-1)\frac{L_{\mathcal S}-1}{L_{\mathcal R}-1}\right),
  \quad
  \mathcal N(\ell)=\operatorname{clip}_{[1,L_{\mathcal S}]}
  \bigl(\pi(\ell)-4,\pi(\ell),\pi(\ell)+4\bigr)
  \label{eq:relative-depth}
\end{equation}
For each role $c\in\{K,V\}$, let
$c_i^{\mathcal M,\ell}\in\mathbb R^{d_{\mathcal M}^{\mathrm{KV}}}$
denote the flattened role-$c$ KV feature at token $i$.
For each aligned token pair $(i_n,j_n)$, we concatenate the selected
sender features:
\begin{equation}
  x_n^{\ell,c}
  =[c_{i_n}^{\mathcal S,t_1}\|c_{i_n}^{\mathcal S,t_2}
  \|c_{i_n}^{\mathcal S,t_3}],
  \quad
  y_n^{\ell,c}=c_{j_n}^{\mathcal R,\ell},
  \quad
  (t_1,t_2,t_3)=\mathcal N(\ell)
  \label{eq:neighborhood-samples}
\end{equation}
where $y_n^{\ell,c}$ is the receiver target. Flattening allows cross-head
mapping even when head counts differ.

\subsection{Moment-Matched Recoloring}
\label{sec:moment-mapping}\label{sec:heterofold}

Token-wise reconstruction objectives~\citep{chen2026dense,linearkv2026} can
preserve low reconstruction error while altering receiver attention and outputs (Figure~\ref{fig:functional-analysis}).
In contrast, Figures~\ref{fig:design-motivation}(b,c) show that Recolor better matches native
K/V distributions than Ridge ($\ell_2$-regularized least squares). Therefore, we match receiver feature means and
covariances through \texttt{Recolor}.

For each receiver layer and role $c\in\{K,V\}$, we collect
$N$ aligned token pairs across the calibration prompts and stack their
features into
$X\in\mathbb R^{N\times\tilde d_{\mathcal S}}$ and
$Y\in\mathbb R^{N\times d_{\mathcal R}}$, where
$\tilde d_{\mathcal S}=3d_{\mathcal S}^{\mathrm{KV}}$ and
$d_{\mathcal R}=d_{\mathcal R}^{\mathrm{KV}}$.
We omit $(\ell,c)$ below when unambiguous and let $x_n$ denote the
$n$th row of $X$. For each layer and role, we compute one scalar RMS
over all sender samples and feature dimensions:
\begin{equation}
r=\sqrt{
\frac{\sum_{n=1}^{N}w_n\lVert x_n\rVert_2^2}
{\tilde d_{\mathcal S}\sum_{n=1}^{N}w_n}
},
\quad
Z=X/r
\label{eq:sender-rms}
\end{equation}
Here, $w_n$ is the sample weight used for moment estimation, and $r$ remains
fixed during inference.
Let $\mu_Z,\mu_Y$, $\Sigma_Z,\Sigma_Y$, and $\Sigma_{ZY}$ denote the
corresponding weighted means, covariances, and cross-covariance.

We whiten both feature spaces to remove their original scales and
correlations, and align the paired features using
Procrustes alignment~\citep{schonemann1966procrustes}.
The thin singular value decomposition gives:
\begin{equation}
\Sigma_Z^{-1/2}\Sigma_{ZY}\Sigma_Y^{-1/2}
=U\Lambda Q^\top,
\quad R^\star=UQ^\top
\label{eq:procrustes}
\end{equation}
After alignment, we restore the receiver covariance and mean to obtain
the \textit{Recolor} initialization:
\begin{equation}
\begin{aligned}
A&=\Sigma_Z^{-1/2}R^\star\Sigma_Y^{1/2},
\quad b=\mu_Y-\mu_ZA, \quad
y^{(0)}=(x/r)A+b
\end{aligned}
\label{eq:heterofold-map}
\end{equation}
For $\tilde d_{\mathcal S}\ge d_{\mathcal R}$ and nonsingular covariances,
the unregularized map satisfies
$\mu_ZA+b=\mu_Y$ and $A^\top\Sigma_ZA=\Sigma_Y$.
Numerical stabilization makes covariance matching approximate in practice.

\subsection{Output-Aware Calibration}
\label{sec:output-aware}

Recolor aligns the cache's global statistical structure but
does not ensure that the transferred cache reproduces native receiver
attention patterns and outputs. 
As illustrated in Figure~\ref{fig:design-motivation}(d), attention-output error persists even after applying Recolor.
To reduce this residual attention-output error, we adapt output-aware
KV calibration for KV quantization~\citep{yun2026optr}
to cross-family cache transfer.
We calibrate K and V separately according to their effects on
receiver computation.

For fixed receiver queries, let $P^*$ and $\widehat P$ denote the native and
transferred attention weights, $V^*$ and $\widehat V$ the corresponding values,
and $W_O$ the receiver output projection. The native and transferred attention
outputs are $O^*=(P^*V^*)W_O^\top$ and
$\widehat O=(\widehat P\widehat V)W_O^\top$.
Their difference can be written as:
\begin{equation}
 \widehat O-O^*
 =\underbrace{[(\widehat P-P^*)V^*]W_O^\top}_{\text{key-induced change}}
 +\underbrace{[\widehat P(\widehat V-V^*)]W_O^\top}_{\text{value-induced change}}
 \label{eq:output-decomposition}
\end{equation}
This separates changes in attention patterns from changes in the retrieved
values, motivating distinct calibration objectives for K and V. For each receiver layer $\ell$ and role $c\in\{K,V\}$, we add a
rank-$\rho_c$ correction to the Recolor output $y_{\ell,c}^{(0)}$ from
Eq.~\eqref{eq:heterofold-map}:
\begin{equation}
  \widehat y_{\ell,c}=y_{\ell,c}^{(0)}
  +(y_{\ell,c}^{(0)}B_{\ell,c})G_{\ell,c},
  \quad
  B_{\ell,c}\in\mathbb R^{d_{\mathcal R}\times\rho_c},\quad
  G_{\ell,c}\in\mathbb R^{\rho_c\times d_{\mathcal R}}
  \label{eq:output-aware-residual}
\end{equation}
The key correction modifies $\widehat P$ in the first term of
Eq.~\eqref{eq:output-decomposition}, while the value correction modifies
$\widehat V$ in the second.

We optimize only $B$ and $G$, with $G$ initialized to zero;
the initial maps and both models remain frozen.
Calibration uses native receiver queries, attention weights, values,
and outputs at tokens overlapping the prompt question span, without gold
answers or generated continuations.

\paragraph{Key and Value Objectives.}
Corrected keys produce $\widehat P$ using native receiver queries,
key normalization, RoPE, and attention scaling.
For query head $h$ and probe $p$, let
$\Delta O^K_{\ell,h,p}
=[(\widehat P_{\ell,h,p}-P^*_{\ell,h,p})V^*_{\ell,g(h)}]W_{O,\ell,h}^\top$
denote the key-induced output change, where $g(h)$ is the corresponding
KV head under grouped-query attention~\citep{ainslie2023gqa}.
Native values remain fixed for the key objective.
For the value objective, we hold the mapped attention weights $\widehat P$
fixed and optimize only the mapped values, forming the full output
$\widehat O_{\ell}(\widehat P,\widehat V)$.

\begin{equation}
\mathcal L_{\ell,K}
=\mathbb E_{x,h,p}\left[
D_{\mathrm{KL}}(P^*_{\ell,h,p}\|\widehat P_{\ell,h,p})
+\frac{\|\Delta O^K_{\ell,h,p}\|_2^2}
{d_{\mathcal R}^{\mathrm{model}}}
\right]
\quad
\mathcal L_{\ell,V}
=\mathbb E_x\left[
\frac{\|\widehat O_{\ell}-O^*_{\ell}\|_F^2}
{\max(\|O^*_{\ell}\|_F^2,\epsilon)}
\right]
\label{eq:output-aware-objectives}
\end{equation}

Here, $d_{\mathcal R}^{\mathrm{model}}$ is the receiver hidden dimension.
The key objective matches attention patterns and their effect after the output
projection, while the value objective refines the values under the fixed
attention pattern without updating the keys. Both losses are averaged across
layers, with equal weight assigned to each example. We optimize them jointly
for four epochs and select the checkpoint with the lowest combined held-out loss.

\paragraph{Folding for inference.}
After calibration, we fold the learned corrections into the initial
affine maps:
\begin{equation}
A^{\mathrm{final}}=A(I+BG),\quad
b^{\mathrm{final}}=b(I+BG),\quad
\Phi_{\ell,c}(x_j)=(x_j/r)A^{\mathrm{final}}+b^{\mathrm{final}}
\label{eq:folded-map}
\end{equation}
Inference uses one fixed affine map per receiver layer and role,
with no separate correction module.
Mapped features are reshaped into receiver KV heads, with receiver
key normalization and RoPE applied to K. Appendix~\ref{app:recolor-details}
gives the optimization settings.

\section{Experimental Results}
\label{sec:experiments}

\begin{table}[t]
\centering
\caption{Cross-family transfer results across six directions. Dense Latent and
KV Ridge use same-index token pairing, while +TA replaces only token
correspondence with our shared character-boundary alignment. TextMas uses
lossless text communication with native receiver prefill. Higher is better.
Bold indicates the best cache-transfer result within each direction and
benchmark; TextMas is excluded from emphasis.}
\vspace{\baselineskip}
\scriptsize
\setlength{\tabcolsep}{1.5pt}
\renewcommand{\arraystretch}{1.05}
\begin{adjustbox}{max width=\textwidth}
\begin{tabular}{@{}ll*{4}{c}@{\hspace{5pt}}*{5}{c}@{}}
\toprule
& & \multicolumn{4}{c}{\textbf{Long Context QA}}
  & \multicolumn{5}{c}{\textbf{Short Context QA}} \\
\cmidrule(lr){3-6}\cmidrule(lr){7-11}
\multirow{2}{*}{\shortstack[l]{\textbf{Transfer}\\\textbf{direction}}}
& \multirow{2}{*}{\textbf{Method}}
& \textbf{Qasper} & \textbf{HotpotQA} & \textbf{LoCoMo} & \textbf{QuALITY}
& \textbf{ARC-C} & \textbf{MMLU} & \textbf{WinoGrande}
& \textbf{HellaSwag} & \textbf{GSM8K} \\
& & (F1 $\uparrow$) & (F1 $\uparrow$) & (F1 $\uparrow$) & (Acc. $\uparrow$)
& (Acc. $\uparrow$) & (Acc. $\uparrow$) & (Acc. $\uparrow$)
& (Acc. $\uparrow$) & (Acc. $\uparrow$) \\
\midrule

\multirow{6}{*}{\shortstack[l]{Qwen3-4B \\ $\rightarrow$ Llama-3.1-8B}}
& TextMas
& 44.64 & 57.54 & 52.33 & 75.07
& 82.76 & 66.74 & 65.82 & 70.78 & 85.67 \\
\cmidrule{2-11}
& Dense Latent
& 2.83 & 0.60 & 0.65 & 23.15
& 24.74 & 24.11 & 50.28 & 48.67 & 0.91 \\
& Dense Latent + TA
& 3.17 & 1.37 & 1.68 & 26.46
& 27.65 & 26.56 & 50.75 & 50.17 & 4.09 \\
& KV Ridge
& 2.78 & 0.74 & 1.41 & 25.17
& 24.40 & 30.36 & 50.36 & 51.40 & 0.53 \\
& KV Ridge + TA
& 16.06 & 34.04 & 13.56 & 54.60
& 76.79 & 61.42 & 52.49 & 59.57 & 43.21 \\
\rowcolor{ourspeach}
& HeteroFold (Ours)
& \textbf{29.12} & \textbf{38.48} & \textbf{29.66} & \textbf{62.85}
& \textbf{78.92} & \textbf{63.28} & \textbf{57.30} & \textbf{65.60} & \textbf{55.12} \\

\midrule
\multirow{6}{*}{\shortstack[l]{Ministral-3-14B \\ $\rightarrow$ Qwen3-4B}}
& TextMas
& 43.07 & 56.05 & 38.92 & 70.61
& 87.80 & 67.95 & 55.09 & 41.54 & 90.14 \\
\cmidrule{2-11}
& Dense Latent
& 2.56 & 0.22 & 0.22 & 24.98
& 25.00 & 27.99 & 51.22 & 44.66 & 1.82 \\
& Dense Latent + TA
& 3.66 & 1.32 & 0.43 & 25.89
& 26.37 & 31.46 & 49.57 & 44.34 & 15.16 \\
& KV Ridge
& 2.77 & 1.45 & 1.73 & 25.41
& 42.15 & 44.27 & 50.99 & 37.45 & 7.05 \\
& KV Ridge + TA
& 39.02 & 48.24 & 14.71 & 73.20
& 88.14 & 71.73 & \textbf{54.78} & \textbf{45.12} & 87.11 \\
\rowcolor{ourspeach}
& HeteroFold (Ours)
& \textbf{42.00} & \textbf{52.74} & \textbf{35.59} & \textbf{75.55}
& \textbf{88.99} & \textbf{72.14} & 53.83 & 41.46 & \textbf{89.23} \\

\midrule
\multirow{6}{*}{\shortstack[l]{Llama-3.1-8B \\ $\rightarrow$ Qwen3-4B}}
& TextMas
& 43.07 & 56.05 & 38.92 & 70.61
& 87.80 & 67.95 & 55.09 & 41.54 & 90.14 \\
\cmidrule{2-11}
& Dense Latent
& 2.72 & 1.40 & 0.67 & 25.31
& 24.49 & 27.00 & 50.67 & 42.96 & 0.76 \\
& Dense Latent + TA
& 4.35 & 1.39 & 0.58 & 25.50
& 25.00 & 27.24 & 51.62 & 43.53 & 6.52 \\
& KV Ridge
& 2.88 & 0.94 & 1.00 & 26.94
& 26.62 & 31.09 & 50.43 & 36.93 & 1.59 \\
& KV Ridge + TA
& 31.63 & 25.12 & 21.91 & 59.59
& 74.32 & 61.25 & 51.38 & \textbf{44.54} & 70.66 \\
\rowcolor{ourspeach}
& HeteroFold (Ours)
& \textbf{35.99} & \textbf{49.44} & \textbf{34.83} & \textbf{62.90}
& \textbf{77.05} & \textbf{62.39} & \textbf{52.88} & 41.74 & \textbf{70.74} \\

\midrule
\multirow{6}{*}{\shortstack[l]{Qwen3-4B \\ $\rightarrow$ Ministral-3-14B}}
& TextMas
& 45.59 & 64.56 & 52.01 & 83.32
& 92.66 & 75.63 & 61.56 & 68.33 & 94.69 \\
\cmidrule{2-11}
& Dense Latent
& 3.19 & 1.66 & 1.23 & 26.51
& 36.26 & 32.07 & 50.04 & 50.62 & 3.87 \\
& Dense Latent + TA
& 5.07 & 4.26 & 3.37 & 38.97
& 42.66 & 41.31 & 50.20 & 50.14 & 28.81 \\
& KV Ridge
& 2.17 & 0.44 & 1.43 & 27.85
& 26.45 & 26.78 & 51.14 & 52.47 & 1.74 \\
& KV Ridge + TA
& 7.75 & 17.09 & 9.33 & 62.94
& 84.13 & 63.41 & 53.12 & 61.74 & 60.20 \\
\rowcolor{ourspeach}
& HeteroFold (Ours)
& \textbf{20.57} & \textbf{42.65} & \textbf{17.75} & \textbf{66.30}
& \textbf{85.49} & \textbf{66.24} & \textbf{55.88} & \textbf{65.22} & \textbf{77.26} \\

\midrule
\multirow{6}{*}{\shortstack[l]{Ministral-3-14B \\ $\rightarrow$ Llama-3.1-8B}}
& TextMas
& 44.64 & 57.54 & 52.33 & 75.07
& 82.76 & 66.74 & 65.82 & 70.78 & 85.67 \\
\cmidrule{2-11}
& Dense Latent
& 2.72 & 0.79 & 0.78 & 24.83
& 31.14 & 28.34 & 51.62 & 51.36 & 1.74 \\
& Dense Latent + TA
& 3.47 & 3.26 & 2.02 & 28.76
& 29.35 & 35.36 & 52.09 & 50.86 & 9.86 \\
& KV Ridge
& 2.67 & 0.54 & 0.26 & 23.54
& 26.45 & 25.62 & 51.62 & 51.61 & 1.67 \\
& KV Ridge + TA
& 19.09 & 37.48 & 39.07 & 74.74
& 88.31 & 68.56 & 52.64 & 65.61 & 53.30 \\
\rowcolor{ourspeach}
& HeteroFold (Ours)
& \textbf{33.72} & \textbf{50.29} & \textbf{43.15} & \textbf{79.00}
& \textbf{90.36} & \textbf{71.76} & \textbf{63.22} & \textbf{69.16} & \textbf{63.46} \\

\midrule
\multirow{6}{*}{\shortstack[l]{Llama-3.1-8B \\ $\rightarrow$ Ministral-3-14B}}
& TextMas
& 45.59 & 64.56 & 52.01 & 83.32
& 92.66 & 75.63 & 61.56 & 68.33 & 94.69 \\
\cmidrule{2-11}
& Dense Latent
& 3.13 & 1.33 & 1.02 & 25.79
& 33.19 & 28.00 & 49.88 & 49.90 & 1.44 \\
& Dense Latent + TA
& 6.71 & 4.88 & 4.43 & 29.72
& 38.57 & 29.58 & 50.51 & 50.41 & 18.65 \\
& KV Ridge
& 1.24 & 0.25 & 0.80 & 23.78
& 22.70 & 26.16 & 51.38 & 48.34 & 0.99 \\
& KV Ridge + TA
& 11.41 & 13.33 & 15.41 & 55.66
& 75.09 & 57.53 & 54.93 & 47.08 & 47.99 \\
\rowcolor{ourspeach}
& HeteroFold (Ours)
& \textbf{21.58} & \textbf{39.00} & \textbf{30.43} & \textbf{72.05}
& \textbf{81.14} & \textbf{65.92} & \textbf{57.14} & \textbf{66.41} & \textbf{72.33} \\

\bottomrule
\end{tabular}
\end{adjustbox}
\label{tab:main_results}
\end{table}

\noindent\textbf{Baselines.}
We evaluate all six directed transfers among Llama-3.1-8B-Instruct~\citep{llama3},
Qwen3-4B~\citep{qwen3}, and Ministral-3-14B-Instruct~\citep{ministral3},
with all models frozen in BF16. TextMas denotes standard text communication. 
In single-hop QA, the sender
passes the original prompt unchanged, and the receiver performs native prefill
to build its own KV cache, providing a text-based reference without cache
transfer or mapping error. 

For Dense Latent~\citep{chen2026dense} and KV Ridge~\citep{linearkv2026},
we reimplement their methods with default mapping and layer-selection settings. Dense
Latent uses proportional layer alignment, while KV Ridge selects $k=8$ sender
layers per receiver layer using calibration $R^2$ and fits separate per-head
K/V maps. Since their original settings assume same-family models with a shared
tokenizer, our direct cross-family variants pair sender token $i$ with receiver
token $i$. The +TA variants replace only this token correspondence with our
shared character-boundary alignment. Full settings are given in
Appendix~\ref{app:baseline-details}.

\noindent\textbf{Tasks.}
We use four long-context QA benchmarks and five short-context QA benchmarks:
\begin{itemize}[leftmargin=*]
    \item Long-context QA: Qasper~\citep{qasper2021}, HotpotQA~\citep{hotpot2018}\thinspace\footnote{We use 200-item
LongBench subsets~\citep{longbench2024} for Qasper and HotpotQA.}, LoCoMo~\citep{locomo2024}, and
QuALITY~\citep{quality2022};
    \item Short-context QA: ARC-Challenge~\citep{arc2018}, MMLU~\citep{mmlu2021}, WinoGrande~\citep{winogrande2020}, HellaSwag~\citep{hellaswag2019}, GSM8K~\citep{gsm8k2021}.
\end{itemize}

We extend our evaluations to multi-round
heterogeneous-agent communication on \textsc{HiddenBench}~\citep{hiddenbench}.
Appendix~\ref{app:settings} lists dataset sizes and decoding settings.
All calibration and benchmark runs are conducted on NVIDIA H100 80GB GPUs.

\noindent\textbf{Calibration setting.}
For all methods, we use same 1,600 training prompts (800 Open-R1~\citep{openr1} and 800 HotpotQA) and 400 held-out
prompts (200 from each), excluding gold answers and solutions.
The HotpotQA prompts used for calibration are drawn from the training split and do not overlap with the evaluation sets.
HeteroFold fits Recolor maps on the training set and selects rank-16
correction checkpoint on the held-out set. KV Ridge uses the training set for
$R^2$-based layer selection and ridge fitting, while Dense Latent uses it for
K/V reconstruction and generates receiver traces for its second stage with a
512-token cap. Appendix~\ref{app:sensitivity} reports
sensitivity to these choices.

\begin{table}[t]
\centering
\caption{Group decisions on \textsc{HiddenBench}.
Initial accuracy precedes communication; the remaining metrics
are measured after 15 rounds.}
\vspace{\baselineskip}
\label{tab:hiddenbench}
\begin{adjustbox}{max width=0.5\textwidth}
\begin{tabular}{@{}lrrrr@{}}
\toprule
\textbf{Communication}
& \begin{tabular}[c]{@{}r@{}}\textbf{Init.}\\\textbf{Acc. (\%)}\end{tabular}
& \begin{tabular}[c]{@{}r@{}}\textbf{Avg.}\\\textbf{Acc. (\%)}\end{tabular}
& \begin{tabular}[c]{@{}r@{}}\textbf{Majority}\\\textbf{Acc. (\%)}\end{tabular}
& \begin{tabular}[c]{@{}r@{}}\textbf{Invalid}\\\textbf{(\%)}\end{tabular} \\
\midrule
TextMas & 11.5 & 29.2 & 27.7 & 4.0 \\
Dense Latent + TA & 11.5 & 18.5 & 7.7 & 3.2 \\
KV Ridge + TA & 11.5 & 21.8 & 16.9 & 20.9 \\
\rowcolor{ourspeach}
HeteroFold (Ours) & 11.5 & \textbf{30.6} & \textbf{29.2} & 3.2 \\
\bottomrule
\end{tabular}
\end{adjustbox}
\vspace{-0.3cm}
\end{table}

\subsection{Results on Cross-Family Transfer}
\label{sec:main-results}

Table~\ref{tab:main_results} compares HeteroFold with direct cross-family
extensions of Dense Latent and KV Ridge, with and without TA. HeteroFold improves over the evaluated cache-transfer
baselines in most settings, with particularly clear gains on long-context QA including unseen benchmarks during calibration,
while avoiding receiver prefill. TextMas provides the native-prefill reference
under lossless text communication.

Giving both baselines the same TA correspondence substantially improves
KV Ridge in many settings, but HeteroFold remains stronger, especially on
long-context tasks. This shows that its gains extend beyond token correspondence
to the K/V mapping and receiver-aware calibration. Additional same-family
comparisons are reported in Appendix~\ref{app:samefamily}.

\subsection{Results on HiddenBench: Multi-Agent Communication}
\label{sec:hiddenbench}

Single-hop QA tests whether cache transfer preserves a fixed prompt, while
multi-agent systems must also communicate newly generated information across
agents. We therefore evaluate HeteroFold on
\textsc{HiddenBench}~\citep{hiddenbench}, where agents must exchange private
information to recover the correct answer. 
Each task uses three or four agents from Llama-3.1-8B, Qwen3-4B, and
Ministral-3-14B. Agents first vote independently, communicate for 15 rounds,
and then vote again.

Unlike single-hop QA, messages are generated during communication. TextMas
sends them as text with native receiver prefill, while Dense Latent, KV Ridge,
and HeteroFold transfer messages through mapped KV caches without
receiver prefill.
HeteroFold achieves performance comparable to TextMas while outperforming the
evaluated cache-transfer baselines. Appendix~\ref{app:hiddenbench-settings}
gives the full evaluation settings.

\subsection{Component Ablation}
\label{sec:component-ablation}

\begin{table}[t]
\centering
\caption{Component ablation for
Ministral-3-14B$\rightarrow$Llama-3.1-8B. ARC-C, GSM8K and QuALITY
report accuracy; Qasper, HotpotQA and LoCoMo report F1. The first three rows replace
token alignment, depth aggregation, and cross-head mixing, respectively. All
variants use the same calibration data as the main results;
Calibration refers to our output-aware calibration.}
\vspace{\baselineskip}
\footnotesize
\begin{adjustbox}{max width=\textwidth}
\begin{tabular}{@{}l*{6}{c}@{}}
\toprule
\textbf{Map} & \textbf{ARC-C} & \textbf{GSM8K} & \textbf{Qasper} &
\textbf{HotpotQA} & \textbf{LoCoMo} & \textbf{QuALITY} \\
\midrule
Same-index token pairing & 69.20 & 7.13 & 3.19 & 0.72 & 1.47 & 24.59 \\
Single-layer mapping & 82.00 & 46.17 & 22.34 & 37.87 & 30.45 & 71.52 \\
Head-local mapping & 71.84 & 26.69 & 9.71 & 17.38 & 12.75 & 54.79 \\
\midrule
TA + Ridge ($\ell_2$-regularized least squares) & 87.03 & 4.55 & 3.27 & 6.82 & 2.13 & 47.41 \\
\quad + key + value calibration
& 89.16 & 55.57 & 21.65 & 31.66 & 31.91 & 74.50 \\
TA + Recolor & 90.19 & 62.77 & 26.52 & 40.71 & 40.81 & 79.34 \\
\quad + key calibration & 90.19 & 62.35 & 34.99 & 48.71 & 42.85 & 78.41 \\
\quad + value calibration & 90.53 & 62.02 & 25.91 & 44.50 & 40.57 & 77.95 \\
\rowcolor{ourspeach}
\quad + key + value calibration (HeteroFold)
& 90.36 & 63.46 & 33.72 & 50.29 & 43.15 & 79.00 \\
\bottomrule
\end{tabular}
\end{adjustbox}
\label{tab:component-ablation}
\end{table}

\begin{table}[t]
\centering
\caption{Transfer latency (in ms, $\downarrow$) and speedup over Native
Prefill ($\times$, $\uparrow$) at batch size~2.
All methods use FlashAttention-2.
Latency sums synchronized stage timings across two
NVIDIA H100 80GB GPUs connected by NVLink, including inter-GPU payload
transfer for cache-transfer methods. Native Prefill processes the original text on the receiver.
Parentheses report speedup over Native Prefill.
Median of three runs after warm-up.}
\vspace{\baselineskip}
\begin{adjustbox}{max width=\textwidth}
\begin{tabular}{@{}l*{3}{r}@{\hspace{8pt}}*{3}{r}@{}}
\toprule
& \multicolumn{3}{c}{\textbf{Llama-3.1-8B $\rightarrow$ Ministral-3-14B}}
& \multicolumn{3}{c}{\textbf{Qwen3-4B $\rightarrow$ Ministral-3-14B}} \\
\cmidrule(lr){2-4}\cmidrule(lr){5-7}
\textbf{Method}
& \textbf{4{,}096 tokens} & \textbf{16{,}384 tokens} & \textbf{32{,}768 tokens}
& \textbf{4{,}096 tokens} & \textbf{16{,}384 tokens} & \textbf{32{,}768 tokens} \\
\midrule
Native Prefill
& 443.1 (1.00$\times$) & 2091.2 (1.00$\times$) & 5167.4 (1.00$\times$)
& 445.1 (1.00$\times$) & 2099.6 (1.00$\times$) & 5178.2 (1.00$\times$) \\

Dense Latent + TA
& 155.4 (2.85$\times$) & 372.1 (5.62$\times$) & 705.7 (7.32$\times$)
& 168.9 (2.64$\times$) & 389.3 (5.39$\times$) & 720.5 (7.19$\times$) \\

KV Ridge + TA
& 135.6 (3.27$\times$) & 301.0 (6.95$\times$) & 569.9 (9.07$\times$)
& 140.4 (3.17$\times$) & 316.5 (6.63$\times$) & 587.7 (8.81$\times$) \\

\rowcolor{ourspeach}
HeteroFold (Ours)
& \textbf{118.3} (3.75$\times$) & \textbf{257.8} (8.11$\times$) & \textbf{481.3} (10.74$\times$)
& \textbf{128.4} (3.47$\times$) & \textbf{290.1} (7.24$\times$) & \textbf{483.2} (10.72$\times$) \\
\bottomrule
\end{tabular}
\end{adjustbox}
\label{tab:efficiency}
\end{table}

Table~\ref{tab:component-ablation} isolates the main components of HeteroFold
for Ministral-3-14B$\rightarrow$Llama-3.1-8B. Replacing TA with same-index
pairing causes the largest degradation, highlighting the importance of
cross-tokenizer alignment. Single-layer and head-local mappings also reduce
performance, supporting multi-layer aggregation and cross-head mixing.
Recolor outperforms Ridge ($\ell_2$-regularized least squares), showing the
benefit of preserving receiver feature statistics. While applying output-aware
calibration substantially recovers Ridge's performance, it still falls short of
TA + Recolor with output-aware calibration, where key correction contributes most of the
gains on long-context tasks.

\section{Analysis}
\begin{figure}[t]
\centering
\includegraphics[width=\linewidth]{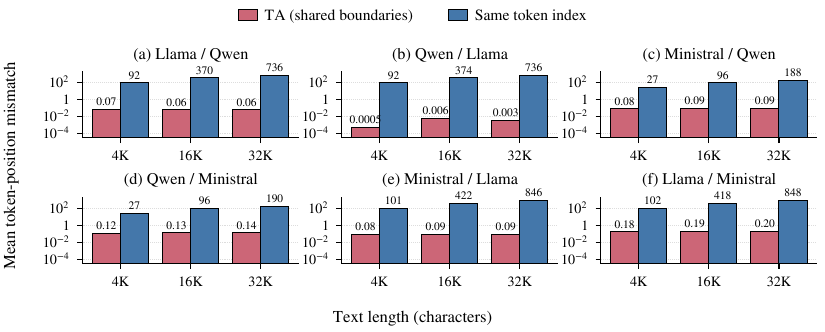}
\caption{Token alignment across all six transfer directions. We compare TA
with same-token-index pairing using the mean absolute difference in character
position between matched sender and receiver tokens. Results use 48
HotpotQA documents at 4K, 16K, and 32K character lengths. Lower is
better. Titles indicate sender / receiver.}
\label{fig:alignment-full}
\end{figure}

\begin{figure}[t]
\centering
\includegraphics[width=\linewidth]{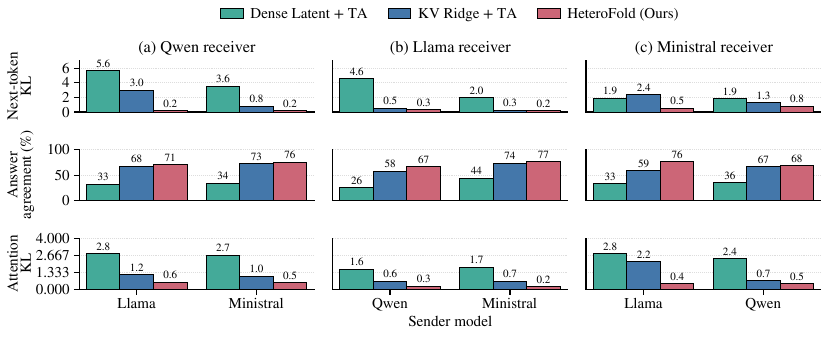}
\caption{Receiver behavior preservation after cache transfer on QuALITY.
Panels (a--c) group results by receiver model and report next-token KL,
answer agreement, and attention KL for the corresponding sender models.
Lower KL and higher agreement indicate closer native receiver behavior.}
\label{fig:threeway-analysis}
\end{figure}
\label{sec:analysis}

\subsection{Latency Measurement}
\label{sec:efficiency}

We measure transfer latency for
Llama-3.1-8B$\rightarrow$Ministral-3-14B and
Qwen3-4B$\rightarrow$Ministral-3-14B on 4K, 16K, and 32K QuALITY
contexts with batch size~2. The sender and receiver reside on two separate
NVIDIA H100 80GB GPUs connected by NVLink. All methods use FlashAttention-2.
Latency sums synchronized timings for
tokenization, TA, inter-GPU payload transfer, K/V mapping, cache construction,
receiver-side processing, and first-token computation. The transferred context
is provided to the receiver as a mapped KV cache without receiver prefill.
Sender prefill, model loading, and offline calibration are excluded. We report
the median of three runs after warm-up. HeteroFold has the lowest latency in
both directions. Compared with Dense Latent's two-layer MLP mapper, HeteroFold
uses a single affine mapping stage. It also uses three sender layers per
receiver layer, compared with eight in KV Ridge. Speedup over Native Prefill
grows from about $3.5$--$3.8\times$ at 4K to about $10.7\times$ at 32K.

\subsection{Cross-Family Transfer Analysis}
\label{sec:AA}

We compare HeteroFold with Dense Latent and KV Ridge, both using TA, while
\emph{Native} denotes direct receiver prefill. Same-index pairing can match
different text positions across tokenizers, and Figure~\ref{fig:alignment-full}
shows that this mismatch grows with context length, while TA maintains close
alignment through shared character boundaries. On QuALITY,
Figure~\ref{fig:threeway-analysis} further shows that HeteroFold more closely
preserves the native receiver's next-token distributions, attention weights,
and answer choices than both baselines across all six transfer directions.
With TA shared across methods, this comparison evaluates their K/V mapping
designs, including HeteroFold's receiver-aware calibration.


\section{Conclusion}
HeteroFold enables prefill-free KV cache transfer across different model
families while keeping both language models frozen. By resolving tokenizer and
model-structure mismatches and calibrating the transferred cache against native
receiver behavior, HeteroFold improves over the evaluated cache-transfer
baselines across six transfer directions, including all four long-context
benchmarks, while reducing receiver-side transfer latency.
These results show that KV computation can be reused across model-family
boundaries without requiring receiver prefill. HeteroFold provides a step
toward efficient cache sharing among heterogeneous language-model agents and
more general KV interfaces across model families.

\label{main-text-end}

\section*{AI Use Statement}

AI tools were used to improve the clarity, grammar, and readability of the manuscript. All AI-assisted edits were reviewed and verified by the authors. The authors take full responsibility for the final content of this work.

\bibliography{iclr2027_conference}
\bibliographystyle{iclr2027_conference}

\clearpage

\appendix
\section*{Appendix: HeteroFold}

\setcounter{figure}{0}
\renewcommand{\thefigure}{\Alph{figure}}
\setcounter{table}{0}
\renewcommand{\thetable}{\Alph{table}}

\makeatletter
\@ifpackageloaded{hyperref}{
  \renewcommand{\theHfigure}{appendix.\Alph{figure}}
  \renewcommand{\theHtable}{appendix.\Alph{table}}
}{}
\makeatother

This appendix provides the evaluation settings, implementation details,
additional receiver analyses, and latency measurements used in the main paper.

\begin{itemize}[leftmargin=*,itemsep=2pt,topsep=3pt]
  \item \textbf{Appendix~\ref{app:settings}.} Experimental Details
  \begin{itemize}[leftmargin=1.5em,itemsep=0pt,topsep=1pt]
    \item \ref{app:models-evaluation} Models and Evaluation
    \item \ref{app:hiddenbench-settings} HiddenBench Evaluation
    \item \ref{app:baseline-details} Calibration and Baseline Implementations
  \end{itemize}

  \item \textbf{Appendix~\ref{app:method-details}.} Alignment and Calibration Details
  \begin{itemize}[leftmargin=1.5em,itemsep=0pt,topsep=1pt]
    \item \ref{app:alignment} TA Implementation
    \item \ref{app:recolor-details} Output-Aware Calibration Details
    \item \ref{app:sensitivity} Sensitivity to Calibration Choices
    \item \ref{app:map-size} Mapper Size and Construction Cost
  \end{itemize}

  \item \textbf{Appendix~\ref{app:samefamily}.} Additional Experimental Results: Same-Family

  \item \textbf{Appendix~\ref{app:diagnostics}.} Additional Receiver Analysis
  \begin{itemize}[leftmargin=1.5em,itemsep=0pt,topsep=1pt]
    \item \ref{app:receiver-behavior} Measuring Receiver Behavior Preservation
    \item \ref{app:reconstruction} Natural-Language Content Preservation
    \item \ref{app:decode-drift} Stability under Autoregressive Decoding
  \end{itemize}

  \item \textbf{Appendix~\ref{app:systems}.} System and Latency Details
  \begin{itemize}[leftmargin=1.5em,itemsep=0pt,topsep=1pt]
    \item \ref{app:latency-measurement} Measurement
    \item \ref{app:latency-breakdown} Latency Breakdown
    \item \ref{app:latency-six-directions} Latency across All Six Directions
    \item \ref{app:sender-cost} Sender-Side Cost
  \end{itemize}
\end{itemize}

\newpage
\section{Experimental Details}
\label{app:settings}

\subsection{Models and Evaluation}
\label{app:models-evaluation}

We use \texttt{meta-llama/Llama-3.1-8B-Instruct},
\texttt{Qwen/Qwen3-4B}, and
\texttt{mistralai/Ministral-3-14B-Instruct-2512-BF16}.
All model weights are frozen. Qwen uses non-thinking mode, while
Ministral uses its language component. Each model tokenizes the same prompt
with its own tokenizer, and native chat-template tokens are handled separately.
We evaluate pairwise cache transfer on short- and long-context QA, together
with multi-round communication on \textsc{HiddenBench}. For the controlled-length
latency table, the sender and receiver are hosted on two separate H100 80GB GPUs
connected by NVLink.

\begin{table}[H]
\caption{Evaluation sets and metrics. Qasper and HotpotQA use LongBench
subsets. LoCoMo excludes unanswerable category 5.}
\vspace{\baselineskip}
\centering
\footnotesize
\begin{tabular}{@{}llrl@{}}
\toprule
\textbf{Suite} & \textbf{Dataset} & \textbf{Examples} & \textbf{Metric} \\
\midrule
Short context
& ARC-Challenge & 1,172 & Accuracy \\
& MMLU & 14,042 & Accuracy \\
& WinoGrande & 1,267 & Accuracy \\
& HellaSwag & 10,042 & Accuracy \\
& GSM8K & 1,319 & Accuracy \\
\midrule
Long context
& Qasper (LongBench) & 200 & Token F1 \\
& HotpotQA (LongBench) & 200 & Token F1 \\
& LoCoMo, categories 1--4 & 1,540 & Token F1 \\
& QuALITY, development & 2,086 & Accuracy \\
\midrule
Multi-agent
& \textsc{HiddenBench} & 65 & Accuracy / invalid rate \\
\bottomrule
\end{tabular}
\label{tab:coverage}
\end{table}

Qasper and HotpotQA use temperature $0.6$, top-$p$ $0.95$, and top-$k$ $20$.
LoCoMo uses greedy decoding with a 64-token generation limit. GSM8K uses
greedy decoding with final-number exact match. HellaSwag and WinoGrande use
length-normalized candidate likelihood, while ARC-Challenge and MMLU use
answer-token likelihood. All methods use the same prompts and decoding
settings within each task.

\subsection{HiddenBench Evaluation}
\label{app:hiddenbench-settings}

We evaluate all 65 Hidden Profile tasks from
\textsc{HiddenBench}~\citep{hiddenbench}. Each participant receives the public
scenario and shared facts, plus one private fact. Three-agent tasks use one
Llama, Qwen, and Ministral participant. Four-agent tasks add a second Qwen
participant. Each task consists of an initial vote, 15 communication rounds,
and a final vote. We use greedy BF16 decoding with limits of 160 tokens per
message and 384 per vote. Invalid votes count as incorrect. We report mean
participant accuracy, majority-vote accuracy, and the invalid-response rate.

TextMas sends generated messages as text for receiver prefill. For cross-family
cache transfer, the sender encodes its message after the public context, and
only the message-span K/V states are mapped to the receiver token grid. Dense
Latent and KV Ridge use their TA variants. Qwen-to-Qwen messages remain text
for every method. Private facts and conversation histories are transferred
only when expressed in a generated message.

\subsection{Calibration and Baseline Implementations}
\label{app:baseline-details}

Calibration uses 1,600 training prompts and 400 held-out prompts, balanced
between Open-R1 problem text and HotpotQA training passages and questions.
Gold answers and solutions are excluded. HeteroFold estimates its moment
statistics on the training split and selects the output-aware correction using
held-out loss.

For single-hop evaluation, TextMas represents lossless text communication.
The original prompt is provided directly to the receiver, which processes it
through native prefill and constructs its own KV cache. In
\textsc{HiddenBench}, the messages are generated during communication, so
TextMas sends each generated message as text and the receiving agent performs
native prefill.

Since there are no public implementations for Dense Latent and KV Ridge, we
reimplement both following their published mapping and layer-selection settings.
KV Ridge uses its default $k=8$ setting, selects sender layers
using training-set head-averaged K/V $R^2$, and fits separate ridge maps with
regularization $0.01$. Dense Latent uses proportional depth routing and
separate two-layer GELU mappings. Its second training stage uses
receiver-generated traces capped at 512 tokens.

Their original evaluations use models within the same family and compatible
tokenization. For the direct cross-family control in
Table~\ref{tab:main_results}, the sender and receiver tokenize the same prompt
separately and sender token $i$ is paired with receiver token $i$. Calibration
stops at the shorter sequence. Receiver positions beyond the sender sequence
reuse the final sender token. The +TA variants change only token correspondence
while preserving each baseline's mapping and layer-selection design.

\section{Alignment and Calibration Details}
\label{app:method-details}

\subsection{Token Alignment (TA) Implementation}
\label{app:alignment}

TA pairs nonempty sender and receiver tokens that end at the same character
position in the original text. A receiver position without an exact match uses
the sender token at the latest preceding shared boundary. Positions before the
first match use the first sender token. In many-to-one cases, sender states are
not averaged; the sender token at the matched character boundary is used.
This forward filling also handles spans that the receiver tokenizes more finely
than the sender. TA is applied to the shared raw-text span, while special and
chat-template tokens are handled separately.
Figure~\ref{fig:alignment-full} measures the resulting position mismatch over
long documents. TA changes only token correspondence. Layer alignment and map
construction remain specific to each method.

\paragraph{Token-position mismatch.}
For each document prefix, we tokenize the same raw text without special tokens
and record the character-end offsets $e_i^{\mathcal S}$ and
$e_j^{\mathcal R}$ of every nonempty sender and receiver token. Let $a(j)$ be
the sender index assigned to receiver token $j$. TA uses the sender token at
the latest shared character boundary, with the first sender token used before
the first shared boundary. Using one-based token indices, same-index pairing
instead uses $a(j)=\min(j,n_{\mathcal S})$, where $n_{\mathcal S}$ is the
sender sequence length. Receiver positions beyond the sender sequence reuse
the final sender token. For the set $J_d$ of nonempty receiver tokens in
document prefix $d$, its mismatch is
\begin{equation}
m_d=\frac{1}{|J_d|}\sum_{j\in J_d}
\left|e_{a(j)}^{\mathcal S}-e_j^{\mathcal R}\right|
\end{equation}
Each bar in Figure~\ref{fig:alignment-full} is the arithmetic mean of $m_d$
over the same 48 documents. Thus every document receives equal weight, and
the quantity is measured in characters. The calculation uses tokenizer offsets
only and does not require a model forward pass.

The three-layer sender neighborhood is clipped at model boundaries. HeteroFold
maps keys before key normalization and RoPE, and values after the value
projection. The receiver then applies its native key normalization and RoPE
once. Mapped K/V features use BF16, while RoPE arithmetic uses FP32.

\subsection{Output-Aware Calibration Details}
\label{app:recolor-details}
\label{app:least-squares}

Moment statistics are estimated independently for K and V at every receiver
layer. The Open-R1 and HotpotQA calibration corpora receive equal total
aligned-token weight. The Recolor initialization matches receiver feature
means and covariances before output-aware calibration.

We evaluate the calibration losses at query positions in the question
span, attending over the full causally accessible prompt.
For the key objective, native receiver queries and values remain fixed.
We match native attention weights through routing KL and compute the
key-induced post-$W_O$ output error separately for each query head,
averaging over heads and question tokens.
For the value objective, we recompute the mapped attention weights
using the current key correction at each training forward pass and
stop gradients through these weights.
We then optimize the mapped values against the native post-$W_O$
output using normalized squared error.
The key and value corrections are optimized jointly, with each
objective updating only its corresponding correction.

The correction rank is $\rho_c=16$. The $B$ factor is initialized from a
zero-mean normal distribution with standard deviation $0.02$, while $G$ starts
at zero. We optimize for four epochs using AdamW with learning rate
$3\times10^{-4}$, weight decay $10^{-4}$, and gradient clipping at 1. We
retain the checkpoint with the lowest combined held-out loss.

\begin{figure}[H]
\centering
\includegraphics[width=\linewidth]{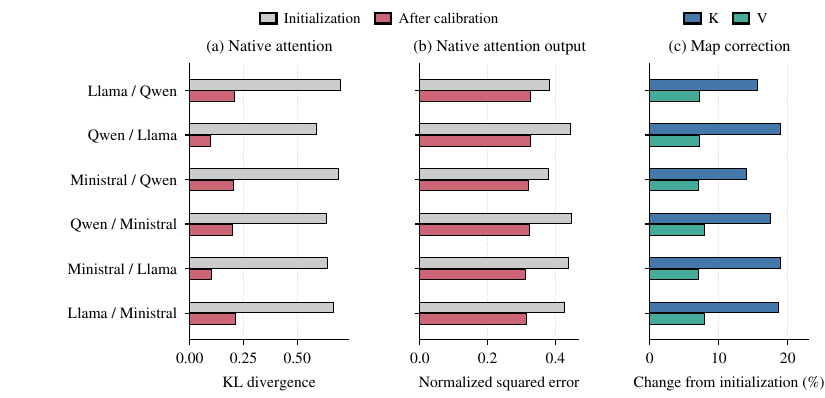}
\caption{Output-aware calibration on held-out prompts. Panels (a,b) compare
the Recolor initialization with the checkpoint selected by the lowest combined
held-out loss using attention KL and normalized attention-output error.
Panel (c) shows the relative change of the affine map. Rows denote
sender / receiver.}
\label{fig:correction-objectives}
\end{figure}

Figure~\ref{fig:correction-objectives} shows that the receiver-aware
calibration objectives decrease across all six transfer directions. The
reported values use the checkpoint selected by the lowest combined held-out
loss. The component ablation in Table~\ref{tab:component-ablation} separately
measures the downstream contribution of Recolor, head mixing, and output-aware
calibration.

The learned low-rank correction is folded into the affine map after
calibration. The inference-time map has the same form as the initial affine
transformation and requires no separate correction module.

\subsection{Sensitivity to Calibration Choices}
\label{app:sensitivity}

Tables~\ref{tab:sensitivity-data}--\ref{tab:sensitivity-corpus} examine the
main calibration choices for Llama-3.1-8B$\rightarrow$Qwen3-4B.
Our default setting uses an equal mixture of Open-R1 and HotpotQA with
1,600 training and 400 held-out prompts, correction rank $\rho_c=16$, and
three sender layers per receiver layer. Each experiment varies one calibration
choice while keeping the remaining settings fixed. The calibration-corpus
comparison uses 800 training and 200 held-out prompts to keep the number of
prompts equal across corpus choices. ARC-C and QuALITY report accuracy, while
HotpotQA reports token F1.

\begin{table}[H]
\caption{Sensitivity to calibration-set size with an equal Open-R1/HotpotQA
mixture, correction rank 16, and three sender layers.}
\label{tab:sensitivity-data}
\vspace{\baselineskip}
\centering
\small
\begin{tabular}{@{}l*{3}{c}@{}}
\toprule
\textbf{Training / held-out prompts}
& \textbf{ARC-C}
& \textbf{HotpotQA}
& \textbf{QuALITY} \\
\midrule
200 / 50 & 72.87 & 26.40 & 61.55 \\
800 / 200 & 75.68 & 41.97 & 63.57 \\
1{,}600 / 400 (paper) & 77.05 & 49.44 & 62.90 \\
\bottomrule
\end{tabular}
\end{table}

Table~\ref{tab:sensitivity-data} shows that increasing the calibration set from
200/50 to 1,600/400 improves ARC-C from 72.87 to 77.05 and HotpotQA from
26.40 to 49.44. QuALITY changes less between the two larger settings.

\begin{table}[H]
\caption{Sensitivity to correction rank with 1,600 training and 400 held-out
prompts from the equal Open-R1/HotpotQA mixture and three sender layers.
Rank 0 uses Recolor without output-aware calibration.}
\vspace{\baselineskip}
\label{tab:sensitivity-rank}
\centering
\small
\begin{tabular}{@{}l*{3}{c}@{}}
\toprule
\textbf{Correction rank $\rho_c$}
& \textbf{ARC-C}
& \textbf{HotpotQA}
& \textbf{QuALITY} \\
\midrule
0 (Recolor only) & 75.43 & 30.47 & 65.10 \\
4 & 75.77 & 47.60 & 63.09 \\
8 & 75.51 & 46.78 & 63.37 \\
16 (paper) & 77.05 & 49.44 & 62.90 \\
32 & 76.71 & 44.45 & 64.57 \\
64 & 76.11 & 43.09 & 64.05 \\
\bottomrule
\end{tabular}
\end{table}

Table~\ref{tab:sensitivity-rank} shows that all nonzero ranks improve HotpotQA
over Recolor alone, while ARC-C and QuALITY vary within a narrower range.
We fix $\rho_c = 16$ for all directions in advance.

\begin{table}[H]
\caption{Sensitivity to the sender-layer neighborhood with 1,600 training and
400 held-out prompts from the equal Open-R1/HotpotQA mixture and correction
rank 16.}
\vspace{\baselineskip}
\label{tab:sensitivity-map}
\centering
\small
\begin{tabular}{@{}l*{3}{c}@{}}
\toprule
\textbf{Sender-layer neighborhood}
& \textbf{ARC-C}
& \textbf{HotpotQA}
& \textbf{QuALITY} \\
\midrule
1 layer & 66.89 & 22.52 & 51.58 \\
3 layers (paper) & 77.05 & 49.44 & 62.90 \\
5 layers & 76.96 & 48.44 & 64.43 \\
\bottomrule
\end{tabular}
\end{table}

Table~\ref{tab:sensitivity-map} shows that using three sender layers improves
ARC-C, HotpotQA, and QuALITY from 66.89/22.52/51.58 to
77.05/49.44/62.90 compared with a single layer. The three- and five-layer
settings give similar results, so we use three layers by default.

\begin{table}[H]
\caption{Sensitivity to the calibration corpus with 800 training and 200
held-out prompts in total, correction rank 16, and three sender layers.
The equal mixture uses 400 training and 100 held-out prompts from each source.}
\vspace{\baselineskip}
\label{tab:sensitivity-corpus}
\centering
\small
\begin{tabular}{@{}l*{3}{c}@{}}
\toprule
\textbf{Calibration corpus (800 / 200)}
& \textbf{ARC-C}
& \textbf{HotpotQA}
& \textbf{QuALITY} \\
\midrule
Open-R1 only & 73.21 & 12.69 & 47.70 \\
HotpotQA only & 74.32 & 36.87 & 60.98 \\
Equal mixture & 75.68 & 41.97 & 63.57 \\
\bottomrule
\end{tabular}
\end{table}

Table~\ref{tab:sensitivity-corpus} shows that the equal Open-R1/HotpotQA
mixture gives the highest result on all three benchmarks, reaching 75.68 on
ARC-C, 41.97 on HotpotQA, and 63.57 on QuALITY. This indicates that combining
reasoning and long-context calibration prompts is more effective than using
either source alone, including on ARC-C and
QuALITY, which are not used for calibration.

\subsection{Mapper Size and Construction Cost}
\label{app:map-size}

HeteroFold stores one folded affine map per receiver layer and role. With three
sender-layer inputs, the mapper contains
\[
2L_{\mathcal R}
\left(
3d_{\mathcal S}^{\mathrm{KV}}d_{\mathcal R}^{\mathrm{KV}}
+d_{\mathcal R}^{\mathrm{KV}}
\right)
\]
parameters. This corresponds to approximately 0.40--0.50\,GB in BF16 for the
evaluated transfer directions. The rank-16 correction is folded into the affine
maps and requires no separate inference parameters. KV Ridge with $k=8$ stores
larger mappings because each receiver layer uses all KV heads from eight
selected sender layers.

\begin{table}[t]
\centering
\caption{Stored mapper parameters per transfer direction, in millions, under
the main experimental settings. HeteroFold uses three sender layers, KV Ridge
uses eight, and Dense Latent uses its two-layer MLP mapper.}
\vspace{\baselineskip}
\label{tab:map-size}
\small
\begin{tabular}{@{}lccc@{}}
\toprule
\textbf{Direction}
& \textbf{Dense Latent}
& \textbf{KV Ridge}
& \textbf{HeteroFold} \\
\midrule
Llama-3.1-8B $\rightarrow$ Qwen3-4B & 303 & 604 & 227 \\
Qwen3-4B $\rightarrow$ Llama-3.1-8B & 270 & 537 & 201 \\
Ministral-3-14B $\rightarrow$ Qwen3-4B & 303 & 604 & 227 \\
Qwen3-4B $\rightarrow$ Ministral-3-14B & 337 & 671 & 252 \\
Ministral-3-14B $\rightarrow$ Llama-3.1-8B & 270 & 537 & 201 \\
Llama-3.1-8B $\rightarrow$ Ministral-3-14B & 337 & 671 & 252 \\
\bottomrule
\end{tabular}
\end{table}

Table~\ref{tab:map-size} shows that HeteroFold stores 201--252 million mapper
parameters across the six transfer directions, approximately 25\% fewer than
Dense Latent and 62\% fewer than KV Ridge. Constructing one HeteroFold transfer
direction takes approximately two H100 GPU-hours under the default calibration
setting of 1,600 training and 400 held-out prompts, three sender layers, and
correction rank 16. This includes moment estimation, receiver-target collection,
and four epochs of low-rank calibration.

\begin{table}[t]
\centering
\caption{Same-family transfer results across four Qwen3 directions.
Dense Latent and KV Ridge follow the token correspondence and mapping settings
of their original same-family formulations, while HeteroFold uses its standard
Token Alignment (TA). All methods use the same calibration data and
train/held-out split as the main experiments. TextMas uses lossless text
communication with native receiver prefill. Higher is better. Bold indicates
the best cache-transfer result within each direction and benchmark; TextMas is
excluded from emphasis.}
\vspace{\baselineskip}
\scriptsize
\setlength{\tabcolsep}{1.5pt}
\renewcommand{\arraystretch}{1.05}
\begin{adjustbox}{max width=\textwidth}
\begin{tabular}{@{}ll*{4}{c}@{\hspace{5pt}}*{5}{c}@{}}
\toprule
& & \multicolumn{4}{c}{\textbf{Long Context QA}}
& \multicolumn{5}{c}{\textbf{Short Context QA}} \\
\cmidrule(lr){3-6}\cmidrule(lr){7-11}
\multirow{2}{*}{\shortstack[l]{\textbf{Transfer}\\\textbf{direction}}}
& \multirow{2}{*}{\textbf{Method}}
& \textbf{Qasper} & \textbf{HotpotQA} & \textbf{LoCoMo} & \textbf{QuALITY}
& \textbf{ARC-C} & \textbf{MMLU} & \textbf{WinoGrande}
& \textbf{HellaSwag} & \textbf{GSM8K} \\
& & (F1 $\uparrow$) & (F1 $\uparrow$) & (F1 $\uparrow$) & (Acc. $\uparrow$)
& (Acc. $\uparrow$) & (Acc. $\uparrow$) & (Acc. $\uparrow$)
& (Acc. $\uparrow$) & (Acc. $\uparrow$) \\
\midrule
\multirow{4}{*}{\shortstack[l]{Qwen3-4B $\rightarrow$ Qwen3-8B}}
& TextMas & 47.81 & 60.34 & 48.18 & 77.09
& 91.47 & 72.35 & 58.72 & 50.79 & 93.40 \\
\cmidrule{2-11}
& Dense Latent & 9.27 & 7.93 & 7.48 & 61.74
& 80.03 & 39.77 & 51.85 & 53.01 & 90.30 \\
& KV Ridge & 35.21 & 50.09 & 34.57 & 69.03
& 85.92 & 67.38 & 54.54 & \textbf{54.56} & 90.22 \\
\rowcolor{ourspeach}
& HeteroFold (Ours) & \textbf{40.27} & \textbf{55.56} & \textbf{42.02} & \textbf{69.27}
& \textbf{86.86} & \textbf{67.89} & \textbf{56.99} & 49.73 & \textbf{90.98} \\
\midrule
\multirow{4}{*}{\shortstack[l]{Qwen3-4B $\rightarrow$ Qwen3-14B}}
& TextMas & 45.70 & 62.50 & 50.42 & 80.63
& 92.32 & 76.56 & 57.93 & 43.23 & 94.01 \\
\cmidrule{2-11}
& Dense Latent & 7.10 & 4.43 & 4.55 & 47.99
& 61.60 & 36.17 & 50.43 & \textbf{52.03} & 38.67 \\
& KV Ridge & 29.02 & 45.16 & 26.05 & 68.46
& \textbf{86.69} & \textbf{67.33} & \textbf{55.25} & 46.42 & 90.22 \\
\rowcolor{ourspeach}
& HeteroFold (Ours) & \textbf{34.13} & \textbf{54.23} & \textbf{37.61} & \textbf{69.70}
& 86.60 & 66.76 & 54.70 & 41.67 & \textbf{90.98} \\
\midrule
\multirow{4}{*}{\shortstack[l]{Qwen3-14B $\rightarrow$ Qwen3-4B}}
& TextMas & 43.07 & 56.05 & 38.92 & 70.61
& 87.80 & 67.95 & 55.09 & 41.54 & 90.14 \\
\cmidrule{2-11}
& Dense Latent & 4.44 & 1.71 & 0.42 & 25.31
& 25.60 & 37.46 & 50.91 & 43.07 & 35.86 \\
& KV Ridge & 39.78 & 44.35 & 35.41 & 72.87
& 90.61 & \textbf{74.03} & \textbf{53.43} & \textbf{44.74} & 89.69 \\
\rowcolor{ourspeach}
& HeteroFold (Ours) & \textbf{41.20} & \textbf{56.06} & \textbf{36.95} & \textbf{75.12}
& \textbf{91.21} & 73.58 & 52.88 & 40.00 & \textbf{91.28} \\
\midrule
\multirow{4}{*}{\shortstack[l]{Qwen3-14B $\rightarrow$ Qwen3-8B}}
& TextMas & 47.81 & 60.34 & 48.18 & 77.09
& 91.47 & 72.35 & 58.72 & 50.79 & 93.40 \\
\cmidrule{2-11}
& Dense Latent & 4.06 & 2.35 & 1.26 & 30.44
& 48.29 & 43.53 & 50.04 & 48.44 & 34.95 \\
& KV Ridge & 36.26 & 53.99 & 39.78 & 76.32
& 89.42 & \textbf{74.31} & 55.33 & \textbf{55.47} & 91.96 \\
\rowcolor{ourspeach}
& HeteroFold (Ours) & \textbf{40.08} & \textbf{55.87} & \textbf{40.63} & \textbf{76.46}
& \textbf{90.61} & 73.86 & \textbf{56.59} & 50.47 & \textbf{92.19} \\
\bottomrule
\end{tabular}
\end{adjustbox}
\label{tab:qwen_same_family}
\end{table}

\section{Additional Experimental Results: Same-Family}
\label{app:samefamily}

To complement the cross-family evaluation, we additionally evaluate HeteroFold
on same-family transfers between Qwen3 models of different scales. For Dense
Latent and KV Ridge, we follow the token correspondence, mapping, and fitting
procedures of their original same-family formulations rather than applying our
Token Alignment (TA). All methods use our calibration setting, including the
same calibration data and train/held-out split as in the main experiments.
These experiments also provide a sanity check for the baseline implementations
in their original same-family transfer setting.

Table~\ref{tab:qwen_same_family} shows that HeteroFold also performs well in
same-family Qwen3 transfers, with consistent gains over the evaluated
cache-transfer baselines on long-context QA. Dense Latent retains reasonable
performance on some short-context tasks but degrades sharply on long-context
QA. KV Ridge remains competitive on several short-context tasks, which also
provides evidence that the baseline implementation behaves as expected in the
same-family setting. HeteroFold remains competitive on short-context QA while
outperforming both baselines on Qasper, HotpotQA, LoCoMo, and QuALITY across
all four transfer directions. These results show that HeteroFold's gains extend
beyond cross-family tokenizer alignment to same-family KV cache transfer.

\section{Additional Receiver Analysis}
\label{app:diagnostics}

\subsection{Measuring Receiver Behavior Preservation}
\label{app:receiver-behavior}

Figure~\ref{fig:threeway-analysis} in the main paper uses all 2,086
QuALITY development questions for each transfer direction. The native receiver
prefills the prompt and serves as the behavioral reference. Each transferred
cache is built from the same sender capture and TA correspondence.

At the answer position, we record the native and transferred receiver
distributions over the full vocabulary, denoted by $p^{\mathrm N}$ and
$p^{\mathrm T}$. For each question, next-token KL is
\begin{equation}
D_{\mathrm{KL}}(p^{\mathrm N}\Vert p^{\mathrm T})
=\sum_v p^{\mathrm N}(v)\log\frac{p^{\mathrm N}(v)}{p^{\mathrm T}(v)}
\end{equation}
The top row of panels~(a--c) reports the median next-token KL across questions
for each direction and method. In the middle row, we restrict both distributions
to the A/B/C/D answer tokens and record whether their highest-probability choices
agree. The displayed value is the percentage of questions with agreement,
measuring behavioral consistency with the native receiver rather than correctness
against the gold answer.

The bottom row of panels~(a--c) evaluates attention preservation on the same
QuALITY questions. The native receiver processes the complete prompt, while the
transferred arm processes the same receiver suffix over its mapped cache. For
every suffix query, receiver head, and layer, let $a^{\mathrm N}$ and $a^{\mathrm T}$
denote the native and transferred attention distributions over the valid key
positions. We compute $D_{\mathrm{KL}}(a^{\mathrm N}\Vert a^{\mathrm T})$ for each
pair and report the mean KL averaged over suffix queries, heads, layers, and
examples.

\subsection{Natural-Language Content Preservation}
\label{app:reconstruction}

To complement the example in Figure~\ref{fig:communication-hero}, we measure
whether a receiver can reconstruct question text available only through the
transferred cache. We sample 100 GSM8K test questions containing 20--60 words.
As in Figure~\ref{fig:communication-hero}, the sender prefills the question,
transfers its KV cache, and the receiver is prompted to restate the question
without access to the original text. For the native reference, the receiver
directly processes the original question tokens; for cache-transfer methods,
the receiver accesses them only through the transferred cache. All methods use
the same Ministral-3-14B$\rightarrow$Llama-3.1-8B direction, BF16, greedy
decoding, and a 192-token generation limit. Table~\ref{tab:question-reconstruction} reports the reconstruction
results.

\begin{table}[H]
\centering
\footnotesize
\caption{Question reconstruction on 100 fixed GSM8K test questions. Ordered
overlap is word-level longest-common-subsequence F1.}
\vspace{\baselineskip}
\begin{tabular}{@{}lccc@{}}
\toprule
\textbf{Method}
& \textbf{Ordered overlap}
& \textbf{Exact}
& \textbf{Reached limit} \\
\midrule
Native receiver & 100.0 & 100/100 & 0/100 \\
HeteroFold & \textbf{83.04} & \textbf{16/100} & 8/100 \\
Dense Latent + TA & 39.74 & 0/100 & 17/100 \\
KV Ridge + TA & 76.68 & 7/100 & 3/100 \\
\bottomrule
\end{tabular}
\label{tab:question-reconstruction}
\end{table}

HeteroFold preserves more of the original wording than the evaluated
cache-transfer baselines. Ordered overlap measures lexical reconstruction
and does not directly measure downstream task accuracy.

\subsection{Stability under Autoregressive Decoding}
\label{app:decode-drift}

Our output-aware calibration uses only prompt tokens, with losses
evaluated at query positions in the question span; no generated
answers or reasoning traces are used for calibration.
We test whether the resulting maps also preserve native receiver
predictions at later decoding positions, beyond those used for
calibration.

For every GSM8K test problem, we first obtain a greedy solution
trace from the native receiver.
Using the frozen calibrated maps, we replay this trace token by
token with either the native prompt cache or a transferred cache.
Both runs receive the same preceding trace tokens at every step.
We compare their next-token distributions using full-vocabulary
KL divergence and top-1 disagreement.
We evaluate Llama-3.1-8B$\rightarrow$Qwen3-4B and
Ministral-3-14B$\rightarrow$Llama-3.1-8B.

\begin{figure}[t]
\centering
\includegraphics[width=\linewidth]{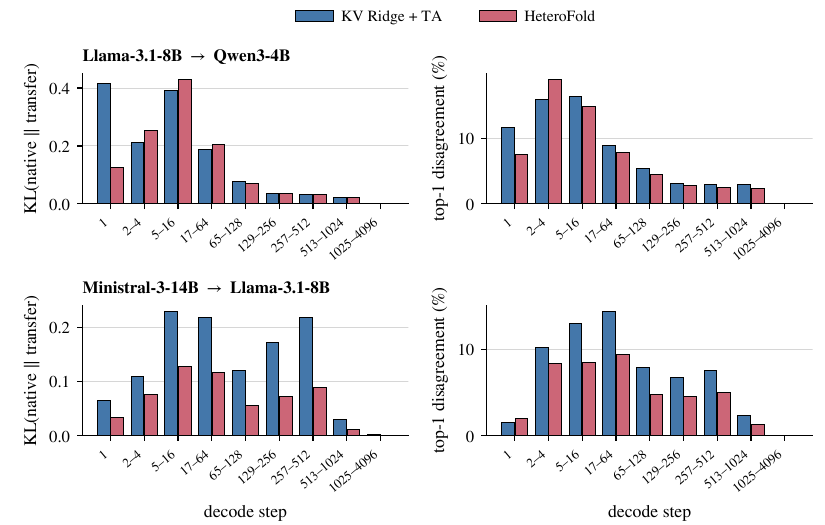}
\caption{Next-token divergence between native and transferred caches along
teacher-forced GSM8K solutions. Left: mean full-vocabulary KL by decode-step
bin. Right: fraction of positions whose top-1 token differs.}
\label{fig:decode-drift}
\end{figure}

Figure~\ref{fig:decode-drift} shows no sustained increase in
prediction divergence with decoding length, with generally smaller
differences at later positions.
These results suggest that prompt-only calibration can preserve
receiver behavior beyond the calibrated query positions.

\newpage

\section{System and Latency Details}
\label{app:systems}

\subsection{Measurement}
\label{app:latency-measurement}

\begin{figure}[t]
\centering
\includegraphics[width=\linewidth]{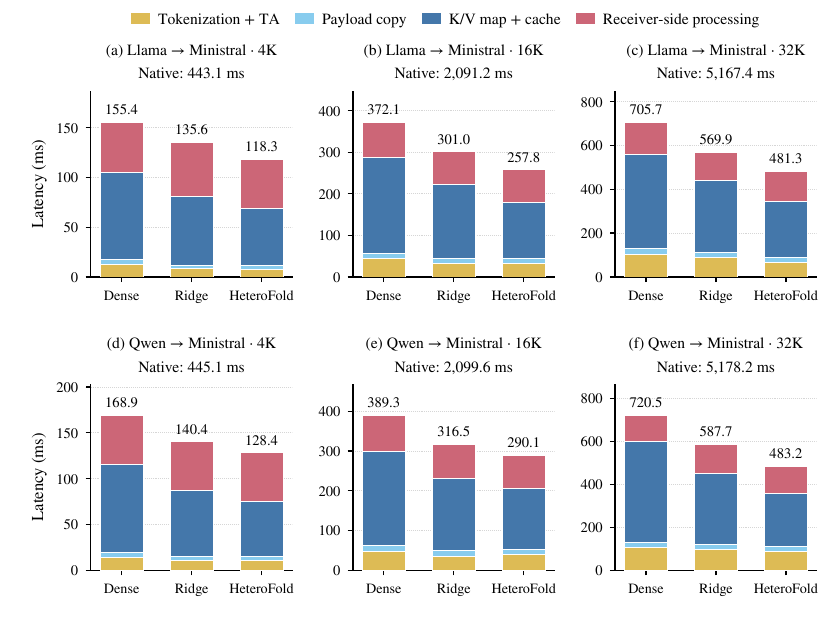}
\caption{Transfer-latency breakdown across two NVIDIA H100 80GB GPUs connected
by NVLink at batch size~2, using FlashAttention-2. The payload stage represents
inter-GPU transfer between the sender and receiver GPUs. The K/V map and
cache stage includes receiver-side cache construction. The final receiver-side
stage processes the mapped cache through the first output token; the transferred
context itself is not re-prefilled by the receiver. Numbers above bars are total
milliseconds. Panel titles give the corresponding Native Prefill time. Each bar
uses the stages from the median-total run among three measurements.}
\label{fig:latency-stages}
\end{figure}

Table~\ref{tab:efficiency} reports the sum of synchronized stage timings
for batches of two QuALITY contexts at 4,096, 16,384, and 32,768
receiver article tokens, before question, options, and chat formatting.
Sender prefill is excluded.

The sender and receiver reside on two separate NVIDIA H100 80GB GPUs in
BF16, connected by NVLink. All methods use FlashAttention-2. Sender K/V
payloads are transferred to the receiver GPU, where K/V mapping, cache
construction, and receiver execution are measured.

For cache-transfer methods, the reported total includes receiver
tokenization, TA, payload copying, K/V mapping, cache construction,
and receiver processing through the first output token for each input.
The transferred context is provided as a mapped KV cache without receiver
prefill. Native Prefill tokenizes and prefills the original text.
Each measured run is preceded by a matched-input warm-up,
and we report the median of three runs. Subsequent decoding is excluded.

Table~\ref{tab:timing-boundary} summarizes the operations included in the measured interval.

\begin{table}[h]
\caption{Operations included in the transfer-latency measurement of
Table~\ref{tab:efficiency}.}
\vspace{\baselineskip}
\centering
\small
\begin{tabular}{@{}lll@{}}
\toprule
\textbf{Stage} & \textbf{Native Prefill} & \textbf{Cache transfer} \\
\midrule
Alignment / mapping & None & TA and K/V map \\
Payload & Original text available & H100-to-H100 K/V copy \\
Transferred context & Native receiver prefill & Mapped KV cache \\
First receiver token & Included & Included \\
Sender prefill & Not required in interval & Excluded \\
Model load / offline fit & Excluded & Excluded \\
Subsequent decoding & Excluded & Excluded \\
\bottomrule
\end{tabular}
\label{tab:timing-boundary}
\end{table}

\begin{figure}[t]
\centering
\includegraphics[width=\linewidth]{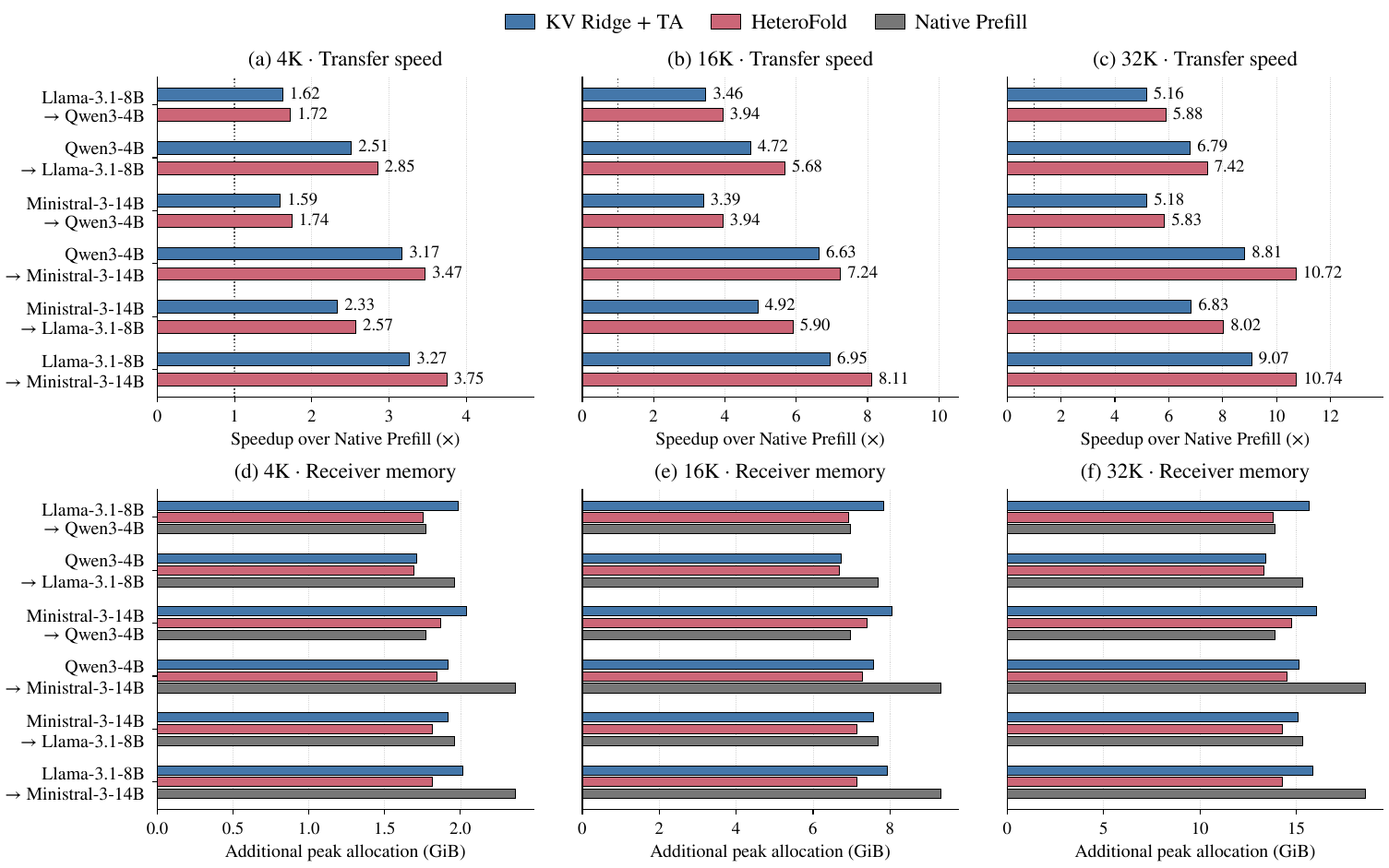}
\caption{Transfer latency and additional receiver memory across all six
directions, using two NVIDIA H100 80GB GPUs connected by NVLink,
BF16 models, FlashAttention-2, and batch size~2.
Sender computation is excluded, while inter-device payload copying is included.
Columns show 4K, 16K, and 32K contexts.
Top: speedup over Native Prefill, computed as the ratio of median latencies
over three runs.
Bottom: median additional peak receiver GPU allocation through the first
output token, relative to the allocation before hand-off. This excludes
resident models, maps, and sender-GPU states, but includes incoming payload
copies, temporary buffers, and receiver caches.
Rows within each panel denote sender $\rightarrow$ receiver.}
\label{fig:latency-six-directions}
\end{figure}

\subsection{Latency Breakdown}
\label{app:latency-breakdown}

Figure~\ref{fig:latency-stages} decomposes the measurements in
Table~\ref{tab:efficiency} for both transfer directions and all three context
lengths. Each bar uses the stages from the run with the median total latency.
Most of HeteroFold's latency advantage comes from its smaller K/V mapping
and cache-construction stage.

\subsection{Latency across All Six Directions}
\label{app:latency-six-directions}

The main latency table focuses on two transfers into Ministral-3-14B at
controlled context lengths and includes an NVLink payload copy.
Figure~\ref{fig:latency-six-directions} extends the comparison to all six
directions at the same 4K, 16K, and 32K context lengths. Each direction
uses two H100 GPUs with batch size~2 and FlashAttention-2.
Sender computation is excluded, while inter-device payload copying is
included. Each configuration is measured three times after matched-input
warm-up, with method order rotated across repetitions.
For the two directions into Ministral-3-14B, the table and both figures
use the same measured runs.

Both transfer methods are faster than Native Prefill in all six directions
at all three context lengths. HeteroFold also has lower latency and uses
less additional peak memory than KV Ridge in every configuration.
With incoming payload copies included, cache transfer does not always
reduce additional peak receiver memory relative to Native Prefill.

\subsection{Sender-Side Cost}
\label{app:sender-cost}

Table~\ref{tab:efficiency} assumes that the sender has already processed the
shared context, as in the intended multi-agent setting. If the sender has not
yet processed the context, sender prefill and payload capture must be added
before cache transfer. For the two-context batches, these costs are
280, 1{,}328, and 3{,}381\,ms for Llama and 235, 1{,}163, and
3{,}083\,ms for Qwen at 4K, 16K, and 32K context lengths, respectively,
when transferring to Ministral. These values include payload-capture
overhead and are reported separately from transfer latency.

\end{document}